\documentclass[11pt]{article}
\usepackage[margin=1in]{geometry}
\usepackage{graphicx}
\usepackage{booktabs}
\usepackage{amsmath,amssymb}
\usepackage{microtype}
\usepackage{authblk}
\usepackage[hidelinks]{hyperref}
\usepackage{caption}
\title{\textbf{Story Operators: Decomposing the Original\,$\to$\,Sequel\\ Transformation in Embedding Space}}
\author{W. Frederick Zimmerman}
\affil{\small Nimble Books LLC \quad \texttt{wfz@nimblebooks.com}}
\date{June 2026}

\begin{document}
\maketitle

\begin{abstract}
\noindent I treat a book as a point in a sentence-embedding space and a literary
transformation as an operation on points. Given an original novel and its sequel, I
ask what it takes, geometrically, to turn the first into the second. Using
all-mpnet-base-v2 paragraph embeddings drawn from a precomputed index of the PG19
corpus, I form the displacement $d=\bar{x}_{\text{seq}}-\bar{x}_{\text{orig}}$ and
greedily decompose it along a content basis obtained by PCA over the two books'
own paragraphs. Each component is an interpretable axis anchored by real passages at
its poles. Across thirteen verified author pairs from Project Gutenberg, the
decomposition reveals a small taxonomy of sequels: \emph{formulaic} (a tiny,
low-rank change: Doyle's Holmes collections, $\lVert d\rVert{=}0.12$),
\emph{concentrated} (one dominant axis: Alcott's \textit{Little Women}\,$\to$\,\textit{Little
Men}, 75\% on a single move), and \emph{compositional} (many small axes:
Twain, Burroughs's Barsoom, Nesbit). For the canonical case,
\textit{Tom Sawyer}\,$\to$\,\textit{Huckleberry Finn}, the dominant recovered axis is
structural---the collapse of sheltering domesticity into a picaresque road---rather
than the famous surface themes of vernacular voice or slavery, which ride later,
smaller axes; and the transformation routes through adventure-journey space rather
than diluting toward generic realism. I corroborate the recovered geometry against
Twain's documented authorial intent (his 1875--76 letters to Howells), which names
the first-person picaresque move years in advance, and I quantify, with an explicit
representation caveat, how much of the realized transformation his stated intentions
span. All computations are reproducible from the released scripts and data.
\end{abstract}

\section{Introduction}
A sentence-embedding model maps text to a vector in $\mathbb{R}^{768}$ in which
proximity encodes semantic kinship. Embedding an entire corpus yields a cloud in
which \textit{Treasure Island} and \textit{Kidnapped} are neighbors and Darwin sits
near the philosophers. Once books are points, transformations of books become
operations on points. I call a small, named algebra of such operations \emph{Story Operators}.
The canonical algebra has 13 primitives \cite{storyops}: 6 \emph{point verbs}
(\textsc{toward}, \textsc{along}, \textsc{scale}, \textsc{reflect}, \textsc{project},
\textsc{blend}), 3 \emph{ensemble verbs} (\textsc{gather}, \textsc{scatter},
\textsc{carry}), and 4 \emph{guards} (\textsc{where}, \textsc{clamp}, \textsc{mask},
\textsc{verify}). The set is argued to be sufficient under an affine-closure analysis:
every affine endomorphism of $\mathbb{R}^d$ decomposes via SVD into rotations,
reflections, axis scalings, and translations, all representable as programs over these
verbs followed by renormalization to $S^{d-1}$ \cite{storyops}. This paper
studies the operator most directly tied to a documented authorial act: the
transformation that carries an \emph{original} novel to its \emph{sequel}. As I
show, the content-axis decomposition of the original$\to$sequel displacement has the
structure of a greedy \textsc{along} chain---\textsc{along} being the canonical verb
for translation along a named semantic direction.

My contribution is threefold. (i) I give a reproducible method that
\emph{decomposes} an original$\to$sequel displacement into a short sequence of
interpretable content axes recovered from the two books' own text, rather than
imposing a fixed external dimension set. (ii) I apply it to thirteen verified
Gutenberg author pairs and identify a taxonomy---formulaic, concentrated,
compositional---supported by comparative metrics. (iii) For
\textit{Tom Sawyer}\,$\to$\,\textit{Huckleberry Finn} I connect the recovered
geometry to primary-source authorial intent and quantify the alignment, with a
forthright account of what the measurement can and cannot claim.

\section{Related Work}
Sentence-BERT \cite{sbert} and the mpnet model family \cite{mpnet} produce the
fixed-length text embeddings I use; I adopt \texttt{all-mpnet-base-v2}. PG19
\cite{pg19} provides a large, public, pre-1919 book corpus with clean provenance.
Prior computational-narrative work largely measures \emph{similarity} or clusters
works; my emphasis is on \emph{decomposing a transformation} into named, composable
operators and validating them against text and, where available, authorial record.
The broader Story Operators program situates these operations in a
reproducing-kernel Hilbert space over narrative corpora \cite{storyops}.

\section{Method}
\subsection{Book vectors}
I use a precomputed paragraph index of PG19 (\texttt{all-mpnet-base-v2}, 768-d,
one vector per paragraph; 12{,}830 distinct books). For a book I discard
boilerplate paragraphs (Project Gutenberg headers/licenses, very short fragments,
illustration captions) and define its vector as the $L_2$-normalized mean of its
remaining paragraph embeddings, $\bar{x}=\mathcal{N}\!\big(\tfrac{1}{|P|}\sum_{p\in P}e_p\big)$.
Paragraph-level pooling avoids the 384-token truncation that would otherwise
discard most of a novel.

\subsection{Content-axis decomposition}
For an (original, sequel) pair with book vectors $T$ and $H$, let
$d=H-T$ and $\mathrm{base}=\langle T,H\rangle$. I build a \emph{content basis}
$\{u_j\}$ by PCA over the union of the two books' centered paragraph embeddings, so
the axes are the directions along which \emph{these two texts actually vary}. I then
greedily compose $d$: order the axes by $|\langle d,u_j\rangle|$, sign-align each
toward $H$, and accumulate $w_k = w_{k-1}+\langle d,u_{(k)}\rangle u_{(k)}$ from
$w_0=T$. After each step I record the cumulative \emph{angular gap closed}
\begin{equation}
g_k=\frac{\cos(w_k,H)-\mathrm{base}}{1-\mathrm{base}},
\end{equation}
and keep steps whose marginal $g_k-g_{k-1}$ exceeds $1\%$ (the others are dead
weight). Each kept step is named from the real paragraphs at the extremes of its
axis (the $\arg\max$/$\arg\min$ of the projection onto $u_{(k)}$), one drawn from the
original and one from the sequel.

In the canonical Story Operators vocabulary \cite{storyops}, each kept step
corresponds to a \textsc{along} operation: $\mathrm{ALONG}(x;\,u,\,\lambda)=\mathcal{N}(x+\lambda u)$,
the canonical verb for translating an embedded object by $\lambda=\langle d,u_{(k)}\rangle$ along
content axis $u_{(k)}$, where $\mathcal{N}$ denotes $L_2$ normalization. Note the
difference from the present method: the canonical verb normalizes after each step,
whereas the greedy decomposition here accumulates in the ambient linear space and
evaluates angular progress, $g_k$, at each step rather than normalizing
incrementally. This gives the decomposition the \emph{conceptual} structure of a
greedy \textsc{along} chain while measuring gap-closure on the original sphere;
the two converge in the limit of small step magnitudes. The greedy ordering
(largest $|\lambda|$ first) recovers the dominant semantic move before the
residual components.

\subsection{Comparative metrics}
Per pair I report: $\cos(T,H)$ (overall similarity); $\lVert d\rVert$
(transformation magnitude); the \emph{content ceiling} $g_K$ (fraction of the
angular gap the content basis closes); $n_{\text{eff}}$ (number of kept steps);
the \emph{dominant share} $g_1$; and a \emph{participation ratio}
$1/\sum_j \hat p_j^2$ over the per-axis energy fractions
$\hat p_j\propto\langle d,u_j\rangle^2$, which is $\approx 1$ when one axis carries
the move and large when the move is diffuse.

\subsection{Intent test (case study)}
For Tom\,$\to$\,Huck I additionally embed six articulated ``sequel operators''
(e.g.\ vernacular first-person POV; town episodes $\to$ river journey; confront
slavery) as phrase directions $\mathrm{dir}=\mathcal{N}(e_{\text{to}}-e_{\text{from}})$
and measure the fraction of $d$ lying in their span, $\lVert B\,d\rVert/\lVert d\rVert$
for an orthonormal basis $B$ of the directions, against a random-$k$-subspace
baseline. I flag the representation caveat in \S\ref{sec:limits}.

\section{Data}
I select thirteen original$\to$sequel pairs whose \emph{both} volumes are present in
the index with $>80$ non-junk paragraphs and which are genuine same-author
continuations (Table~\ref{tab:comp}). They span children's fantasy (Carroll, Baum,
Nesbit), boyhood and pulp adventure (Twain, Burroughs), detective (Doyle),
historical romance (Dumas), domestic (Alcott, Montgomery, Porter), and utopian
(Bellamy) modes. Gutenberg identifiers and per-book paragraph counts are recorded in
the released \texttt{comparative.csv}.

\section{Results}
\subsection{A taxonomy of sequels}
\begin{figure}[t]\centering
\includegraphics[width=\textwidth]{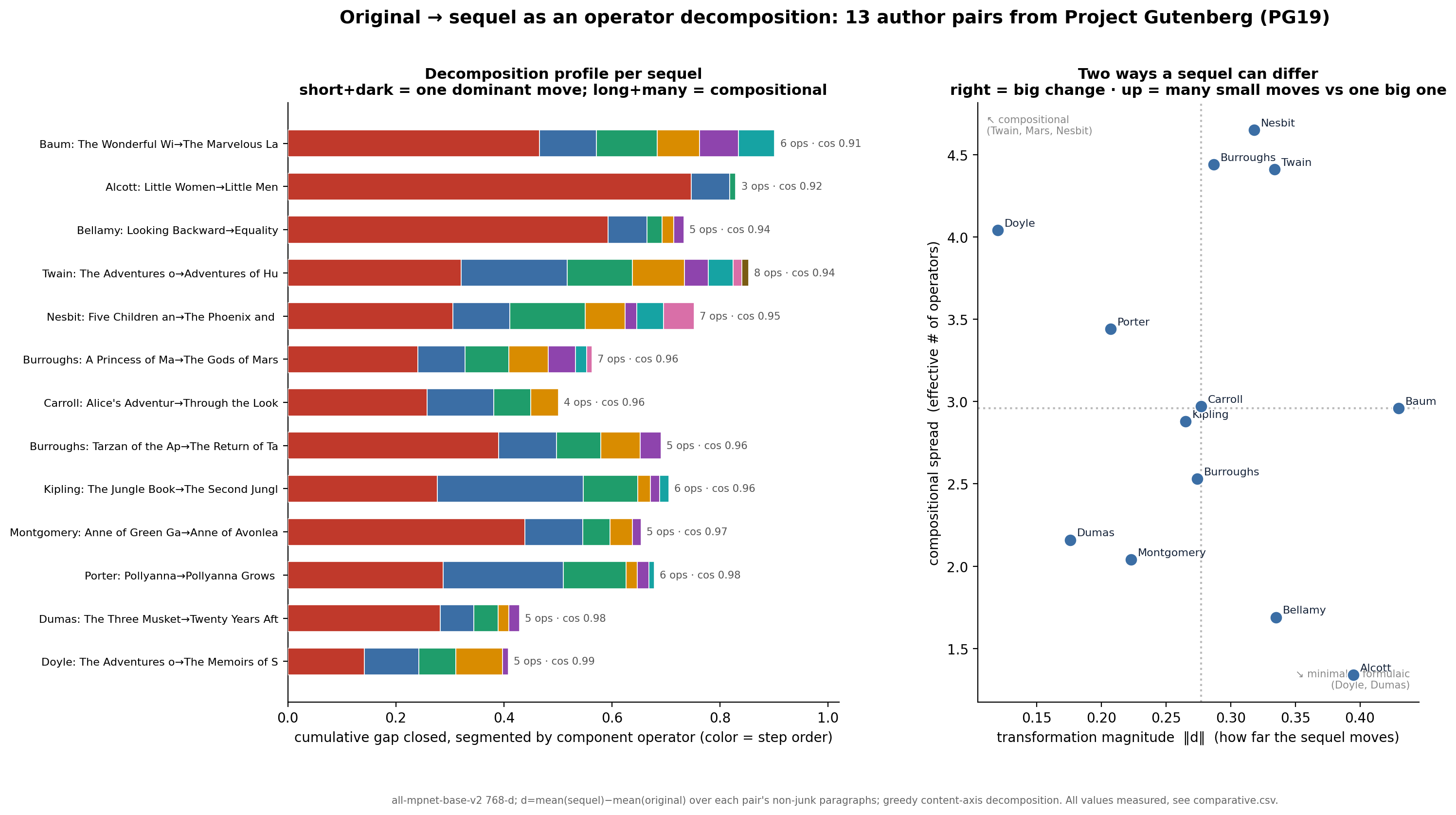}
\caption{Comparative profiles. \textit{Left}: cumulative gap closed per sequel,
segmented by component operator (color = greedy step order); short-and-dark bars are
single-dominant-axis sequels, long-and-many are compositional. \textit{Right}:
transformation magnitude $\lVert d\rVert$ vs.\ compositional spread (participation),
separating formulaic (lower-left) from compositional (upper) sequels.}
\label{fig:comp}
\end{figure}

Table~\ref{tab:comp} and Fig.~\ref{fig:comp} show that sequels differ along two
nearly independent axes: \emph{how far} the sequel moves ($\lVert d\rVert$) and
\emph{how concentrated} the move is (participation).
\begin{itemize}
\item \textbf{Formulaic.} Doyle's \textit{Adventures}\,$\to$\,\textit{Memoirs of
Sherlock Holmes} is the least-transformed pair ($\cos=0.99$, $\lVert d\rVert=0.12$,
dominant axis only 14\%): the detective template barely changes between story
collections. Dumas's \textit{Three Musketeers}\,$\to$\,\textit{Twenty Years After}
is similar ($\lVert d\rVert=0.18$).
\item \textbf{Concentrated.} Alcott's \textit{Little Women}\,$\to$\,\textit{Little
Men} puts 75\% of the move on a single axis (participation 1.3): the shift from a
girls' domestic coming-of-age to a boys' schoolhouse is nearly one operator.
Bellamy's \textit{Looking Backward}\,$\to$\,\textit{Equality} is comparable (59\%),
the utopian narrative hardening into didactic treatise.
\item \textbf{Compositional.} Twain, Burroughs's
\textit{Princess}\,$\to$\,\textit{Gods of Mars}, and Nesbit spread the move across
many small axes (participation 4.4--4.7): the sequel is a genuine re-composition,
not a single tonal turn.
\end{itemize}
Baum's \textit{Wizard}\,$\to$\,\textit{Marvelous Land of Oz} is the
\emph{most}-transformed pair ($\lVert d\rVert=0.43$, ceiling 90\%), consistent with
Baum dropping Dorothy for an entirely new protagonist and cast.

\begin{table}[t]\centering\small
\caption{Original$\to$sequel decomposition across thirteen Project Gutenberg pairs,
sorted by transformation magnitude. $\cos$: original--sequel cosine; $\lVert
d\rVert$: displacement norm; \emph{ceil}: content ceiling (gap closed); \emph{stp}:
effective steps; \emph{dom}: dominant-axis share; \emph{part}: participation ratio.
All values measured.}
\label{tab:comp}
\begin{tabular}{@{}p{2.6cm}p{5.0cm}rrrrrr@{}}
\toprule
Author & Original $\to$ Sequel & $\cos$ & $\lVert d\rVert$ & ceil & stp & dom & part\\
\midrule
L. Frank Baum & The Wonderful Wizard of Oz $\to$ The Marvelous Land of Oz & 0.91 & 0.43 & 90\% & 6 & 47\% & 3.0 \\
Louisa May Alcott & Little Women $\to$ Little Men & 0.92 & 0.40 & 84\% & 3 & 75\% & 1.3 \\
Edward Bellamy & Looking Backward $\to$ Equality & 0.94 & 0.34 & 75\% & 5 & 59\% & 1.7 \\
Mark Twain & The Adventures of Tom Sawyer $\to$ Adventures of Huckleberry Finn & 0.94 & 0.33 & 85\% & 8 & 32\% & 4.4 \\
E. Nesbit & Five Children and It $\to$ The Phoenix and the Carpet & 0.95 & 0.32 & 76\% & 7 & 31\% & 4.7 \\
Edgar Rice Burroughs & A Princess of Mars $\to$ The Gods of Mars & 0.96 & 0.29 & 57\% & 7 & 24\% & 4.4 \\
Lewis Carroll & Alice's Adventures in Wonderland $\to$ Through the Looking-Glass & 0.96 & 0.28 & 52\% & 4 & 26\% & 3.0 \\
Edgar Rice Burroughs & Tarzan of the Apes $\to$ The Return of Tarzan & 0.96 & 0.27 & 71\% & 5 & 39\% & 2.5 \\
Rudyard Kipling & The Jungle Book $\to$ The Second Jungle Book & 0.96 & 0.27 & 71\% & 6 & 28\% & 2.9 \\
L. M. Montgomery & Anne of Green Gables $\to$ Anne of Avonlea & 0.97 & 0.22 & 66\% & 5 & 44\% & 2.0 \\
Eleanor H. Porter & Pollyanna $\to$ Pollyanna Grows Up & 0.98 & 0.21 & 69\% & 6 & 29\% & 3.4 \\
Alexandre Dumas & The Three Musketeers $\to$ Twenty Years After & 0.98 & 0.18 & 44\% & 5 & 28\% & 2.2 \\
Arthur Conan Doyle & The Adventures of Sherlock Holmes $\to$ The Memoirs of Sherlock Holmes & 0.99 & 0.12 & 42\% & 5 & 14\% & 4.0 \\
\bottomrule
\end{tabular}
\end{table}

\subsection{Case study: \textit{Tom Sawyer}\,$\to$\,\textit{Huckleberry Finn}}
\begin{figure}[t]\centering
\includegraphics[width=\textwidth]{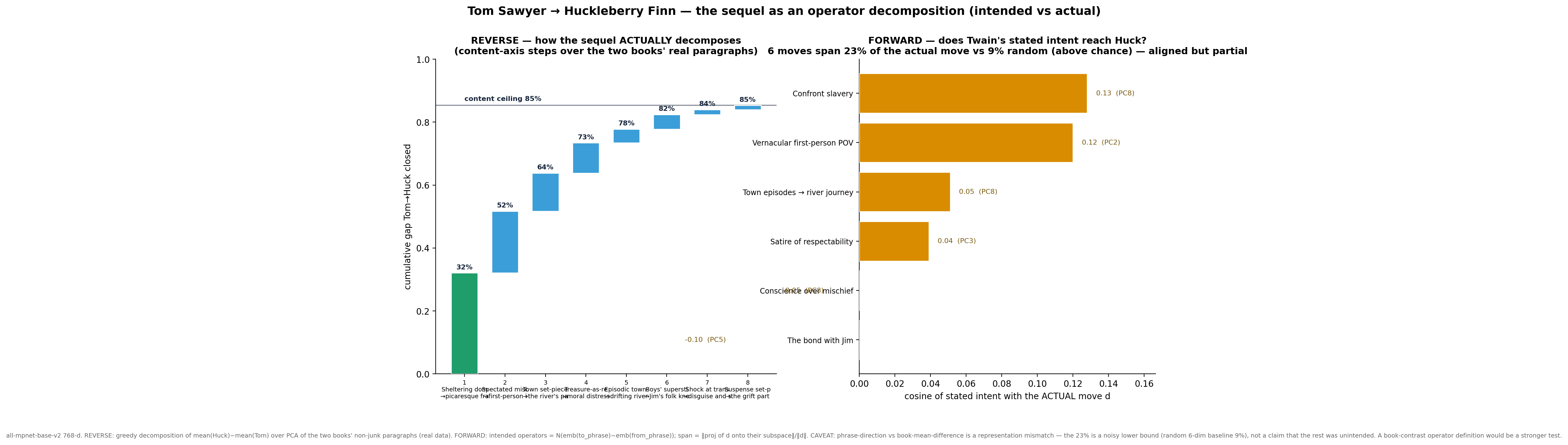}
\caption{Tom\,$\to$\,Huck. \textit{Left}: the actual decomposition (cumulative gap by
content-axis step; content ceiling 85\%). \textit{Right}: forward intent test---six
articulated authorial moves span 22.8\% of the actual displacement vs.\ an 8.9\%
random-6-dim baseline (above chance, modest), with the representation caveat noted in
\S\ref{sec:limits}.}
\label{fig:reverse}
\end{figure}

\begin{figure}[t]\centering
\includegraphics[width=\textwidth]{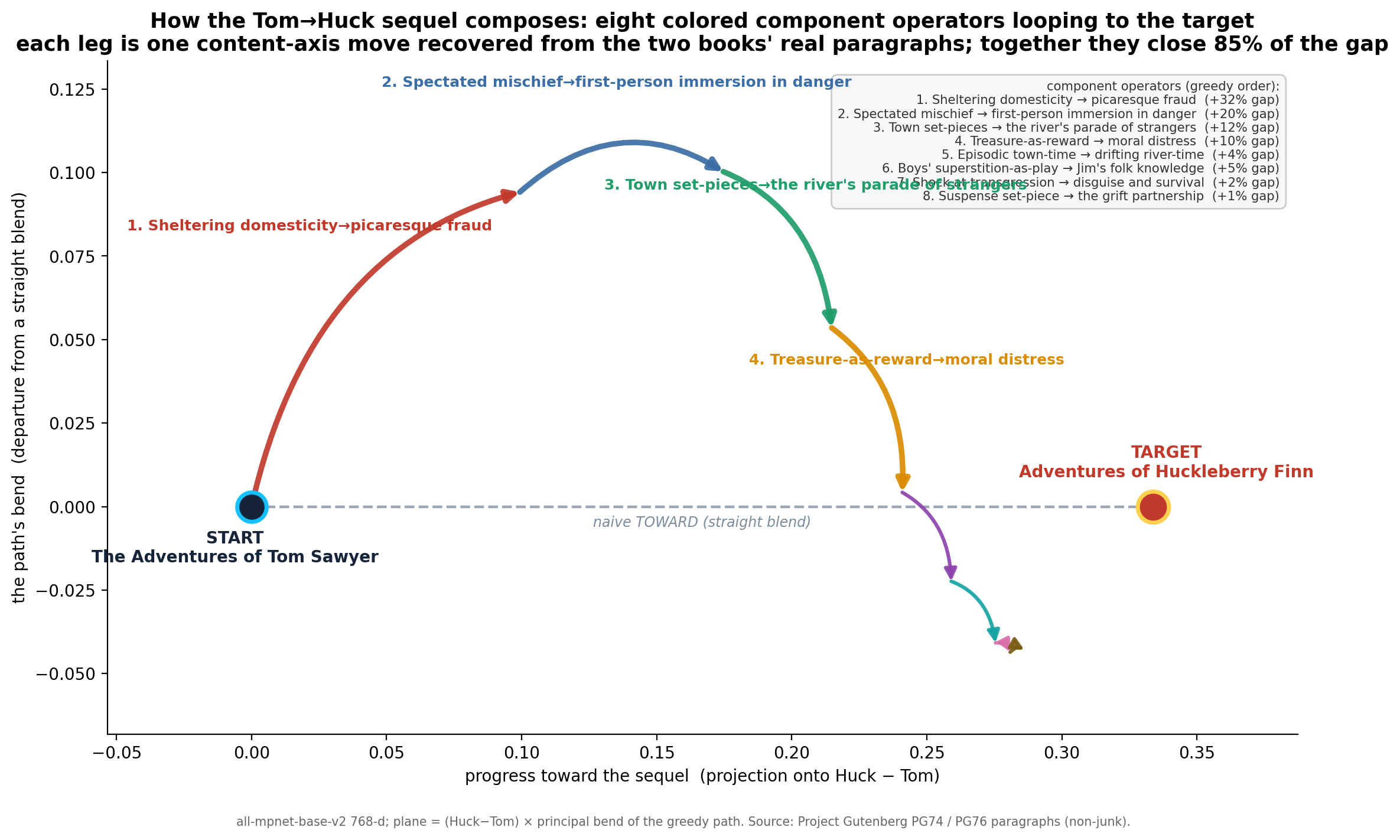}
\caption{The eight component operators as colored legs looping from
\textit{Tom Sawyer} (start) to \textit{Huckleberry Finn} (target) in the plane
spanned by $H-T$ and the path's principal bend. The dashed line is the naive straight
blend; the legend lists each operator's marginal gap.}
\label{fig:components}
\end{figure}

The Twain pair is strongly compositional (8 effective steps, participation 4.4,
ceiling 85\%). The four dominant axes (74\% of the move), each named from real
passages, are: (1, 32\%) \emph{sheltering domesticity}~$\to$~\emph{picaresque fraud}
(Aunt Polly praying over Tom $\to$ the con-man ``king''); (2, 20\%) \emph{spectated
mischief}~$\to$~\emph{first-person immersion in danger}; (3, 12\%) \emph{town
set-pieces}~$\to$~\emph{the river's parade of strangers}; (4, 10\%)
\emph{treasure-as-reward}~$\to$~\emph{moral distress}. The single largest axis is thus
\emph{structural}---the protective adult frame collapsing into a road of
frauds---rather than the surface themes (vernacular voice, slavery) usually named
first; those ride later, thinner axes. Tracing the bent path's neighbors, the
transformation \emph{routes through} adventure-journey space (\textit{Kidnapped},
\textit{Treasure Island} appear at the 25--50\% marks), whereas the naive straight
interpolation merely dilutes Tom toward generic social realism
(Figs.~\ref{fig:reverse},~\ref{fig:components}).

\subsection{Recovered geometry vs.\ authorial intent}
The recovered dominant axes match Twain's documented plan. Nine years before he
finished \textit{Huckleberry Finn}, Clemens wrote to W.\,D.\ Howells (5 July 1875)
that he had stopped \textit{Tom Sawyer} at boyhood and intended, separately, to
``\,take a boy of twelve \& run him on through life (in the first person) \dots\ but
not Tom Sawyer\,'' \cite{howells}---naming the first-person picaresque move (axes 1--2)
in advance, and invoking the picaresque \textit{Gil Blas} as the model. The 9 August
1876 letter dates the sequel's start (``It is Huck Finn's Autobiography\,\dots\ I like
it only tolerably well\,\dots\ may possibly pigeon-hole or burn the MS''); the
published \textsc{notice} and \textsc{explanatory} note assert, respectively, the
absence of conventional plot (consistent with the ``drifting river-time'' axis) and
the deliberate, researched vernacular. Twain's own later verdict on the book---a
notebook entry (c.\ 1895) describing ``a sound heart \& a deformed conscience [that]
came into collision \& conscience suffers defeat'' \cite{conscience}---names the
moral-distress axis (4).

Quantitatively, embedding the six stated intentions as phrase directions, they span
$22.8\%$ of the realized displacement, versus an $8.9\%$ random-6-dimensional
baseline (95th percentile $12.9\%$): above chance but partial. The dominant
structural axis (shelter $\to$ fraud) is the \emph{least} matched by any stated
intention---suggesting that the deepest move of the sequel was less a checklist item
than an emergent consequence of following Huck honestly down the river.

\subsection{Other notable pairs}
Beyond the extremes, several pairs illustrate intermediate regimes. Carroll's
\textit{Alice}\,$\to$\,\textit{Through the Looking-Glass} has the lowest content
ceiling (52\%): much of its change lies off the two books' shared content axes,
consistent with a sequel that swaps one self-contained dream-logic for another
rather than continuously transforming a single world. Montgomery's \textit{Anne of
Green Gables}\,$\to$\,\textit{Anne of Avonlea} is the most nearly two-dimensional
(participation 2.0): a steady maturation along essentially one pair of axes.
Kipling's two \textit{Jungle Book}s, Burroughs's two Tarzan novels, and Porter's two
\textit{Pollyanna} books sit in the middle---moderate magnitude, moderate
spread---the profile of a continuation that keeps its world but advances its
protagonist. That these regimes recur across unrelated authors suggests the taxonomy
reflects modes of sequel-making rather than idiosyncrasies of single works.

\section{Discussion and Limitations}\label{sec:limits}
\textbf{Two kinds of difference.} Magnitude and concentration are nearly orthogonal:
a sequel can be a large but single-axis turn (Baum, Alcott) or a small but diffuse
re-weighting (Doyle). The taxonomy is descriptive, but it cleanly separates
template continuations from genuine re-compositions.

\textbf{Caveats.} (i) \emph{Representation mismatch in the intent test}: intended
operators are phrase-embedding differences while $d$ is a book-paragraph-mean
difference; the 22.8\% figure is therefore a noisy \emph{lower bound} on
intent-alignment, not a claim that the remainder was unintended. A book-contrast
definition of each operator (centroids of real exemplar books) would be a stronger
test. (ii) The content basis is unsupervised PCA over two books; it captures the
texts' own variance, not a canonical narrative axis set. (iii) Mean-pooling discards
sequence; the operators describe aggregate semantic content, not plot order.
(iv) The angular ``gap closed'' and the Euclidean $\lVert d\rVert$ measure different
geometry (a residual can be long yet nearly orthogonal to the target). (v) Thirteen
pairs is a small, English, pre-1919 sample. None of these undercut the central, fully
reproducible result---that an original$\to$sequel transformation decomposes into a
few interpretable, text-anchored operators---but they bound its interpretation.

\textbf{Applications.} Decomposed transformations are useful beyond criticism. The
same geometry that names a sequel's moves can score a \emph{proposed} continuation
against its original---does it re-compose the world or merely dilute it?---place a
catalog's series in the magnitude--concentration plane to diagnose sameness, and
supply retrieval-time structure for curation and recommendation. Within the canonical
13-primitive algebra \cite{storyops}, this paper is consistent with the utility of
\textsc{along}: the sequel cases confirm that named semantic translations (content-axis
projections) compose into interpretable programs and that the dominant move is
recoverable from text alone. I use the full operator algebra in this role within a
working publishing program; the present study provides empirical grounding for the core
vocabulary on a case where ground-truth authorial record exists, before relying on it where it does not.

\textbf{Reproducibility.} Every number here is produced by released scripts
(\texttt{sequel\_corpus\_decompose.py}, \texttt{sequel\_figures.py},
\texttt{tom\_huck\_*}) over the public PG19 texts; the comparative table is generated
directly from the computed \texttt{comparative.json}.

\section{Conclusion}
Treating books as points and transformations as operators turns ``what does this
sequel do?'' into a measurable question. Across thirteen author pairs the
original$\to$sequel move decomposes into a handful of interpretable, text-anchored
axes, and sequels fall into recognizable kinds. For \textit{Tom
Sawyer}\,$\to$\,\textit{Huckleberry Finn}, the recovered geometry both matches Twain's
stated intent on the first-person picaresque turn and exposes a deeper structural
move---sheltering domesticity collapsing into a picaresque road---that no stated
intention fully anticipated. Story Operators thus offer a vocabulary in which literary
transformation is not only describable but decomposable.

\end{document}